\documentclass[a4paper, 10pt, twocolumn]{article}

\usepackage[utf8]{inputenc}
\usepackage{amsfonts,amssymb,amsmath,amsthm}
\usepackage{graphicx}
\usepackage[footnotesize]{caption}
\usepackage{subcaption}
\usepackage{hyperref}
\usepackage{bm}
\usepackage{algorithm}
\usepackage{algorithmic}

\usepackage[top=1.2cm, left=1.2cm, right=1.2cm, bottom=1.2cm]{geometry}

\title{Off-the-grid data-driven optimization of sampling schemes in MRI}

\author{Alban Gossard$^{1,2}$, Frédéric de Gournay$^{1,2}$ and Pierre Weiss$^{1,2}$.\\
\footnotesize $^1$Institut de Mathématiques de Toulouse.\
$^2$Institut des Technologies Avancées du Vivant. Université de Toulouse \& INSA  \& CNRS.
}
\date{\empty} % no need for a date

\renewenvironment{abstract}{\bf\small {\em\ Abstract---}}{}

\DeclareMathOperator*{\argmin}{\mathop{\mathrm{argmin}}}

\def\R{\mathbb{R}}
\def\C{\mathbb{C}}
\def\dist{\mathrm{dist}}
\def\A{\mathcal{A}}
\def\M{\mathcal{M}}
\def\x{\mathbf{x}}

\begin{document}

\maketitle

\begin{abstract}
We propose a novel learning based algorithm to generate efficient and physically plausible sampling patterns in MRI. This method has a few advantages compared to recent learning based approaches: i) it works off-the-grid and ii) allows to handle arbitrary physical constraints. These two features allow for much more versatility in the sampling patterns that can take advantage of all the degrees of freedom offered by an MRI scanner. The method consists in a high dimensional optimization of a cost function defined implicitly by an algorithm. We propose various numerical tools to address this numerical challenge.
\end{abstract}

\vspace{-0.3cm}
\section{Introduction}
\label{sec:introduction}

The design of efficient sampling patterns in MRI is a critical issue with a long history \cite{bernstein2004handbook} and a renewed interest in recent years with the advent of compressed sensing and deep learning.
\paragraph{State-of-the-art} The most recent trends can be separated in two families
\begin{description}
 \item[\emph{Compressed sensing theory:}] In a recent set of works, the theory of compressed sensing was improved to more closely fit the practical issues of MRI \cite{chauffert2014variable,adcock2017breaking,boyer2016generation,boyer2019compressed}. In a nutshell, these works suggest that good sampling schemes should have a variable density: the low frequencies should be sampled more densely than the high frequencies (though this has little to do with the quantity of energy present in the signal) and the samples should cover the space locally uniformly. This led a few authors to generate sampling schemes by minimizing the distance bewteen a measure $\mu$ belonging to a set of admissible measures and a continuous target probability density function $\pi$:
 \begin{equation}\label{eq:optimize_measure}
  \inf_{\mu\in \A} \dist(\mu,\pi),
 \end{equation}
 where $\dist$ is a distance that metrizes the weak convergence, such as a discrepancy \cite{graf2012quadrature,chauffert2017projection} or the Wasserstein distance \cite{lebrat2019optimal} and $\A$ is a set of admissible probability measures such as the set of discrete measures supported on $M$ points $\M_M=\{\mu=\frac{1}{m}\sum_{m=1}^M \delta_{x_m}, x_m\in \R^d\}$ or more exotic sets of contraints that more closely describe the physical constraints of a scanner. This approach led to remarkable practical results \cite{lazarus2019sparkling} that are currently evaluated for clinical routine.

 \item[\emph{Learning:}] Motivated by the recent breakthroughs of learning and deep learning, many authors recently tried to \emph{learn} either the reconstructor \cite{hammernik2018learning}, the sampling pattern \cite{gozcu2018learning,sherry2019learning}, or both \cite{jin2019self}. In \cite{gozcu2018learning}, the authors propose a greedy algorithm that generates a sampling pattern by iteratively selecting a discrete horizontal line that maximizes the SNR of the reconstructed image. A similar principle is proposed in \cite{jin2019self}, but there, the reconstructor is learnt simultaneously. In \cite{sherry2019learning}, the authors adopt a similar approach, but replace the greedy algorithm by a bi-level programming approach that controls the number of sampling points using an $\ell^1$ penalization. Overall, all those works suffer from the same limitations:
 \begin{itemize}
  \item The sampling points are required to live on a Cartesian grid, which is suboptimal.
  \item The methods cannot incorporate advanced constraints on the sampling trajectory and therefore focus on ``rigid'' constraints such as imposing to sample horizontal lines.
  \item The methods are computationally intensive, which may be not be so critical since the sampling schemes are generated offline. %but still raise the question of carbon footprint \cite{strubell2019energy}.
 \end{itemize}
 To the best of our knowledge, the only paper that addresses the above criticisms is the just posted \cite{weiss2019pilot}, where the authors simultaneously optimize a reconstructor and a sampling scheme by performing a local optimization of a well initialized trajectory.
\end{description}

\paragraph{Our contribution} In this paper, we propose to blend both approaches by using a data-driven distance in \eqref{eq:optimize_measure} rather than more principled approaches. This allows us to avoid all the above-mentioned flaws. This also leads us to implement a set of advanced numerical routines to address the computational challenges raised by the proposed cost function.

\vspace{-0.1cm}
\section{The proposed approach}

\subsection{Preliminaries}
We assume that both a set of training images $\x=(x_1,\hdots,x_K)\in \C^{N\times K}$ and a differentiable image quality metric $\eta:\R^N\times \R^N\to \R_+$ are available.
In this work, we will simply consider the squared $\ell^2$ distance $\eta(\hat x,x)=\frac{1}{2}\|\hat x - x\|_2^2$. In what follows, we let $\xi \in (\R^d)^M$ denote a set of locations in the $k$-space (or Fourier domain), $y\in \C^M$ denote a set of $k$-space measurements and $R:\C^M\times (\R^d)^M \to \C^N$ denote a fixed reconstructor, i.e. for a sampling scheme $\xi\in (\R^d)^M$ and a measurement vector $y\in \C^M$, we let $\hat x = R(\xi, y)$ denote the reconstructed image. We let $A(\xi)\in \C^{M\times N}$ denote the forward Fourier transform defined for all $1\leq m\leq M$ and $x\in \C^N$
\begin{equation}
 [A(\xi) x]_m = \sum_{n=1}^N x_n \exp(-i \langle p_n, \xi_m\rangle),
\end{equation}
where $p_n\in \{-N/2,\hdots, N/2-1\}^d$ are the positions of grid points in the image space. For a regularization parameter $\lambda >0$,
we consider a Tikhonov reconstructor:
\begin{equation}\label{eq:reconstructor_tikhonov}
 R_1(\xi, y)=\argmin_{x\in \C^N} \frac{1}{2}\|A(\xi)x - y\|_2^2 + \frac{\lambda}{2}\|x\|_2^2,
\end{equation}
and a nonlinear compressed sensing type reconstructor
\begin{equation}\label{eq:reconstructor_nonlinear}
 R_2(\xi, y)=\Psi z^\star, \quad z^\star=\argmin_{z\in \C^P} \frac{1}{2}\|A(\xi)\Psi z - y\|_2^2 + \lambda \|z\|_1,  
\end{equation}
where $\Psi\in \C^{N\times P}$ is a redundant wavelet transform.

\subsection{The principle}
\label{sec:first-section}

The goal here is to replace the distance in \eqref{eq:optimize_measure} by a data-driven cost function. A natural choice reads:
\begin{equation}\label{eq:main_problem}
\min_{\xi \in \Xi} F(\xi) := \mathbb{E}\left( \sum_{x\in \x} \eta(R(\xi, A(\xi)x + b) , x) \right), \tag{$\mathcal{P}$}
\end{equation}
where $b\sim \mathcal{N}(0,\sigma^2 I_N)$ and $\Xi\subseteq (\R^d)^M$ describes the physical constraints. In words, the term $A(\xi)x + b$ represents noisy data acquisition that we want to reconstruct as well as possible, in average, using the reconstructor $R$. The expectation is taken w.r.t. the noise realizations.

\paragraph{Differentiating the reconstructors}
Solving \eqref{eq:main_problem} is a real computational challenge. It is high dimensional, the cost function does not have a simple analytic formula and its regularity properties are unclear. Since the cost function is defined through another minimization problem,  \eqref{eq:main_problem} can be interpreted as a bi-level optimization problem. Various approaches are available to solve it \cite{bard2013practical}. Here we will follow the approach suggested in \cite{ochs2015bilevel}. Instead of solving the lower-level minimization problems \eqref{eq:reconstructor_tikhonov} or \eqref{eq:reconstructor_nonlinear} exactly, we assume that they are solved approximately using iterative algorithms such as a conjugate gradient method or a proximal gradient descent. The main idea is then to differentiate the algorithm using dedicated libraries such as PyTorch instead of the minimizer itself.

\paragraph{Implementing and differentiating the NUFT} The fast implementation of the linear mapping $A(\xi)$ is the backbone of our approach. It corresponds to the non uniform Fourier transform (NUFT). Various efficient approximate implementations have been devised over the past \cite{fessler2003nonuniform,greengard2004accelerating,keiner2009using} and Python toolboxes begin to emerge \cite{lin2018python}. Our experience using them however led to unstable results due to significant numerical errors. In this work, we therefore opted for a direct (naive) implementation of the NUFT on massively parallel architectures, following the numerical experiments conducted in \href{https://www.kernel-operations.io/keops/index.html}{KeOps}. The main observation is that for a GPU with 1TFlop, applying the NUFT to small $128\times 128$ images (which is typical in this field) just requires a fraction of second, which is compatible with large scale computations. We therefore implemented a homemade NUFT within PyTorch, allowing for automatic differentiation.

\paragraph{Optimizing the cost function} The previous details allow to automatically compute the derivative of $F$ w.r.t. $\xi$ when replacing the expectation by an empirical average. This in turn allows to use any off-the-shelf optimization solver. In this preliminary work, we simply set $\Xi=(\R^d)^M$ (i.e. no constraints between samples), and $b=0$ (no noise in the measurements) and used a limited memory BFGS algorithm. More advanced stochastic gradient approaches are expected to be used later.

\section{Results}
\label{sec:results}

Here we report preliminary results with this approach.
Two $64\times 64$ images are studied and we compare 3 patterns sub-sampled at a factor $3.3$. Our approach is abbreviated OSP (optimal sampling pattern), we also use a low-frequency pattern (LF) and a variable density sampler with a uniform density (VDS). The images are reconstructed both with the linear \eqref{eq:reconstructor_tikhonov} and nonlinear reconstructors \eqref{eq:reconstructor_nonlinear}.
Without surprise, LF is good at reconstructing global shape of images and removing noise but the VDS performs better to reconstruct details (with the disadvantage of generating noise). Our OSP combines both advantages.
Our approach shows that choosing an optimized $k$-space improves the peak noise-to-signal ratio (PSNR) between $2$dB and $10$dB for these test-cases in comparison to variable density sampling or standard low-frequency sampling.
In the future, we plan to focus on learning a reconstructor, adding physical constraints to the set $\Xi$ and introducing the noise $b$.

\begin{figure}[t!]
    \centering
    \begin{subfigure}[t]{0.125\textwidth}
        \centering
        \includegraphics[width=\textwidth]{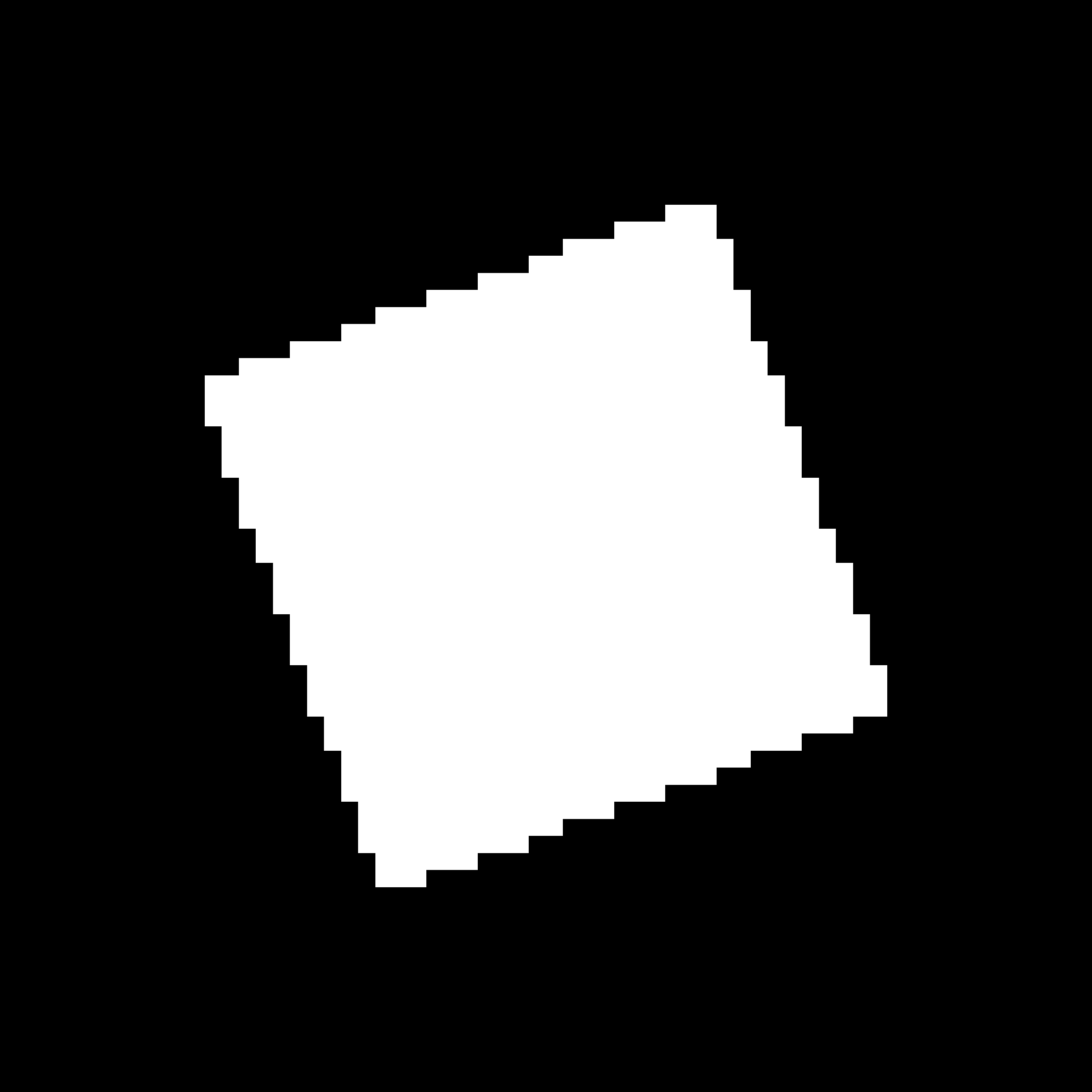}
        \caption{ }
        \label{fig:orig_square_img}
    \end{subfigure}%
    \begin{subfigure}[t]{0.125\textwidth}
        \centering
        \includegraphics[width=\textwidth]{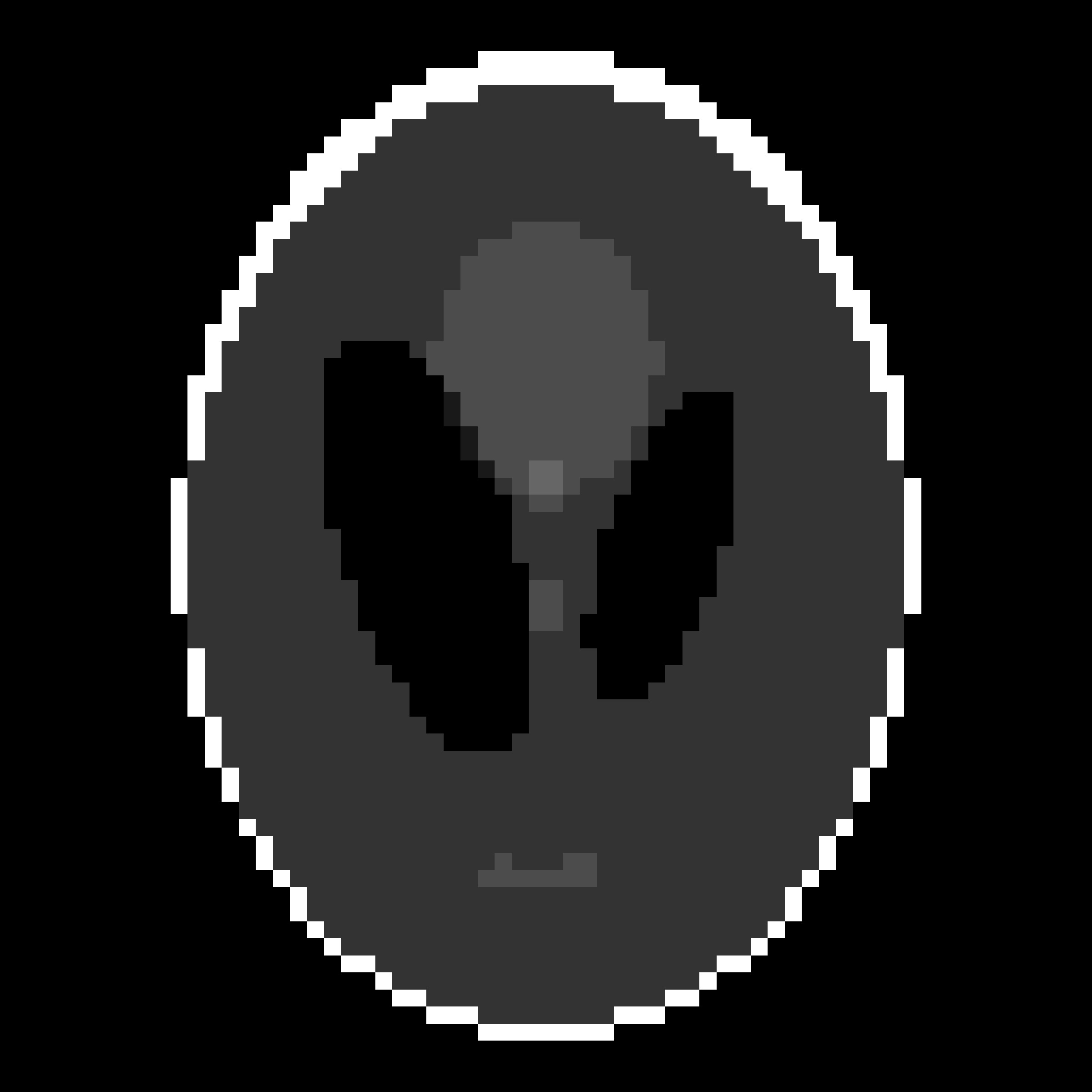}
        \caption{ }
        \label{fig:orig_phantom_img}
    \end{subfigure}%
    \begin{subfigure}[t]{0.125\textwidth}
        \centering
        \includegraphics[width=\textwidth]{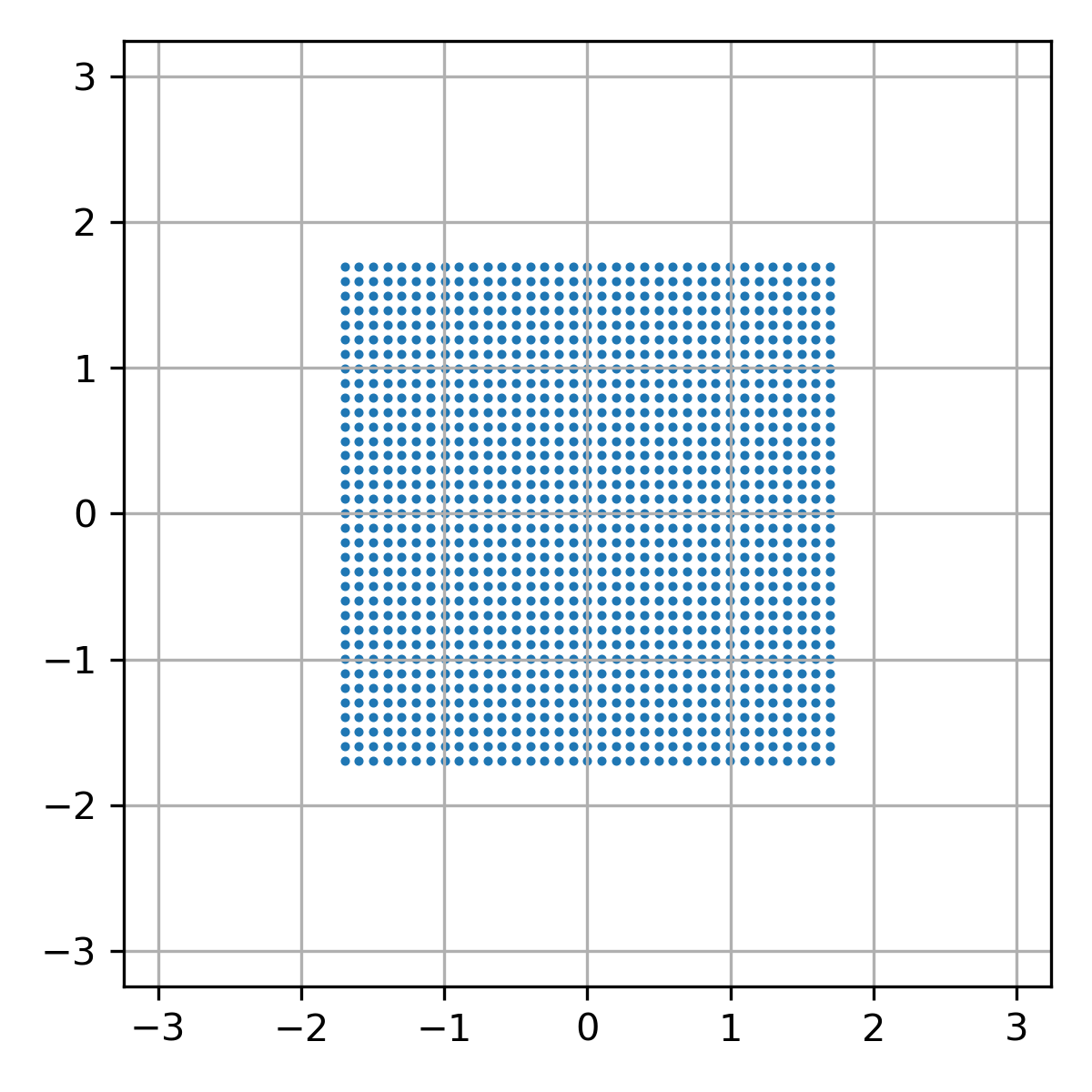}
        \caption{ }
        \label{fig:xi_BF}
    \end{subfigure}%
    \begin{subfigure}[t]{0.125\textwidth}
        \centering
        \includegraphics[width=\textwidth]{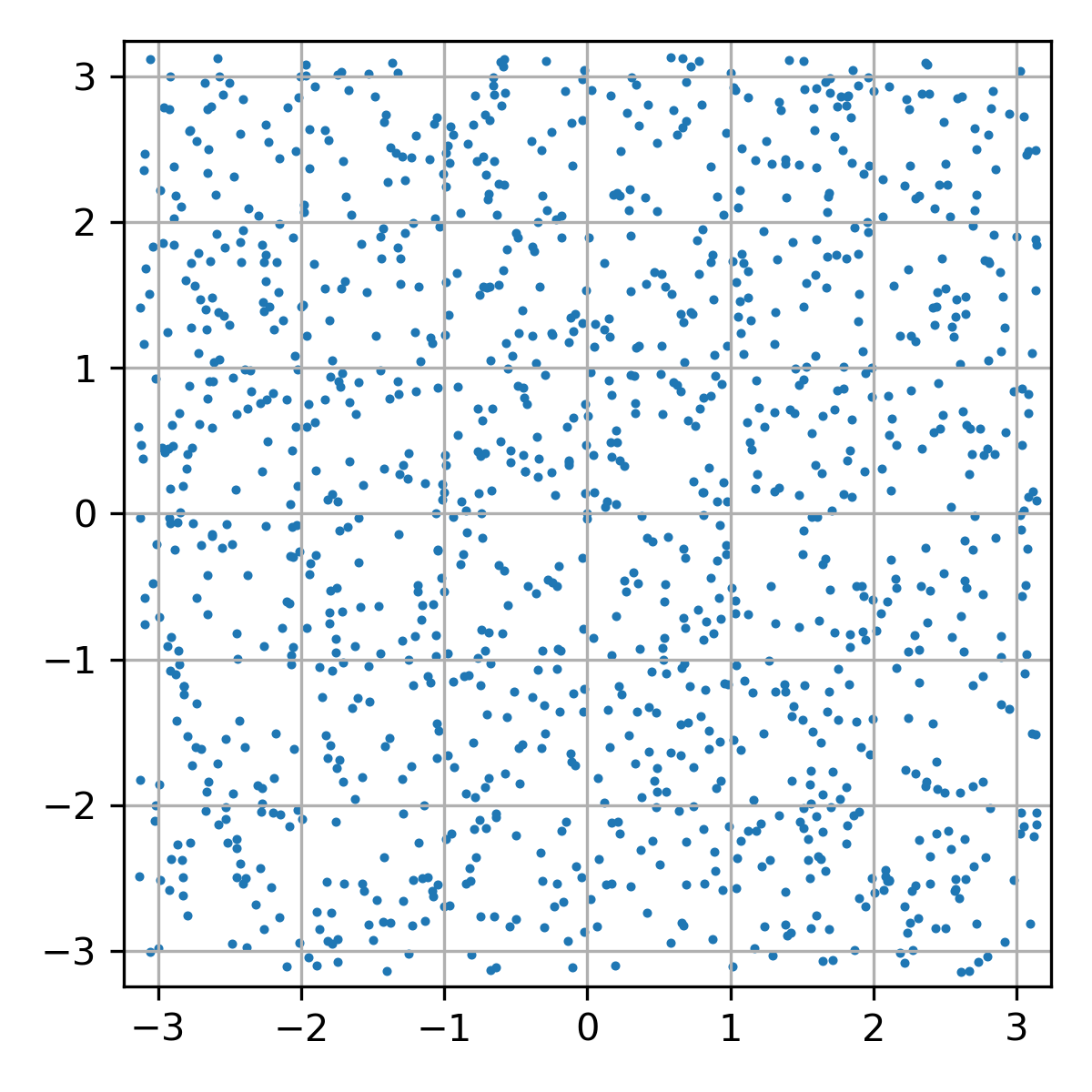}
        \caption{ }
        \label{fig:xi_DB}
    \end{subfigure}
    \caption{Original images (\ref{fig:orig_square_img}) square and (\ref{fig:orig_phantom_img}) phantom. Different $k$-space sampling pattern: (\ref{fig:xi_DB})VDS and (\ref{fig:xi_BF}) LF random sampling.}
\end{figure}

\begin{figure}[t!]
    \centering
    \begin{subfigure}[t]{0.125\textwidth}
        \centering
        \includegraphics[width=\textwidth]{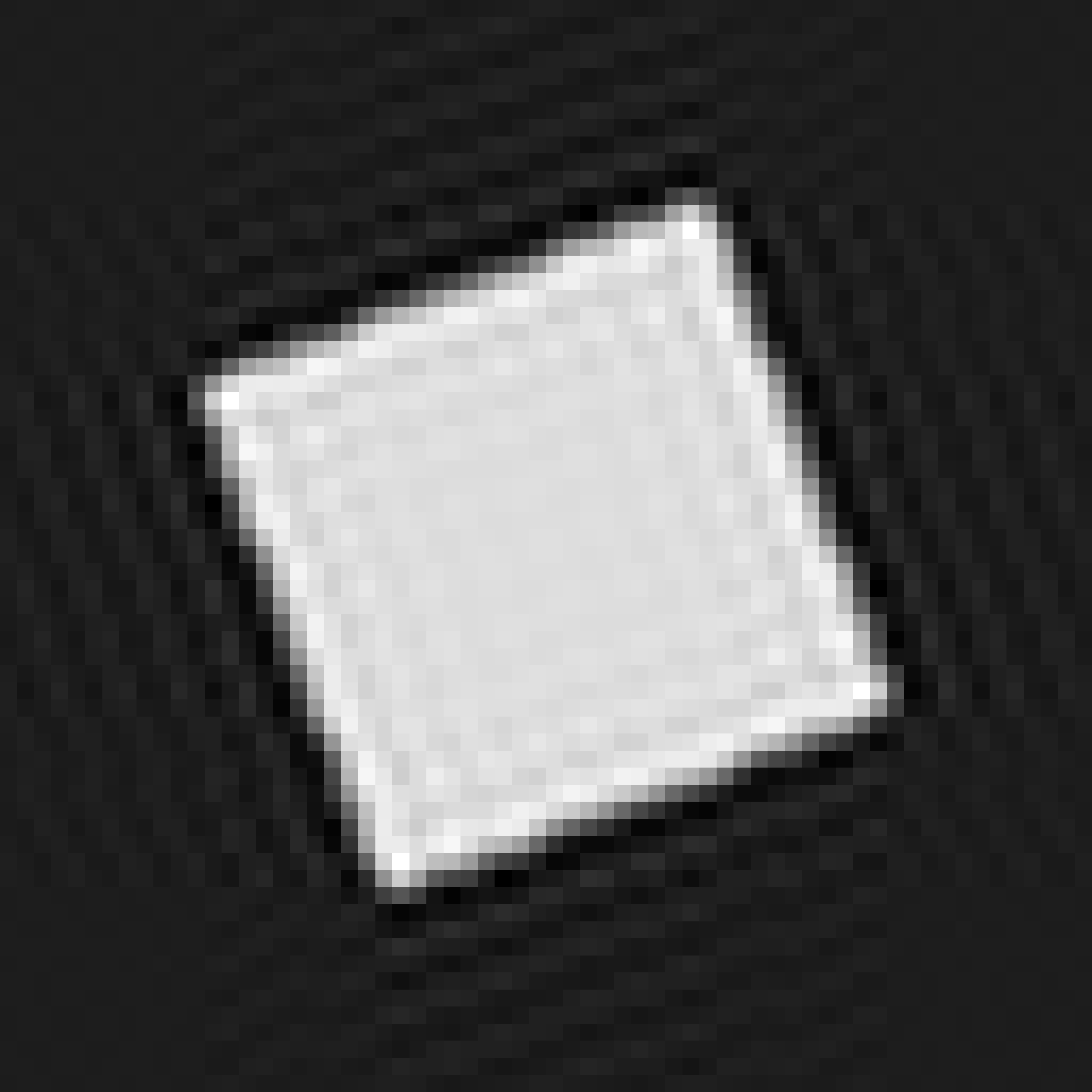}
        \caption{PSNR$=53.3$dB}
        \label{fig:square_BF_tikhonov}
    \end{subfigure}%
    \begin{subfigure}[t]{0.125\textwidth}
        \centering
        \includegraphics[width=\textwidth]{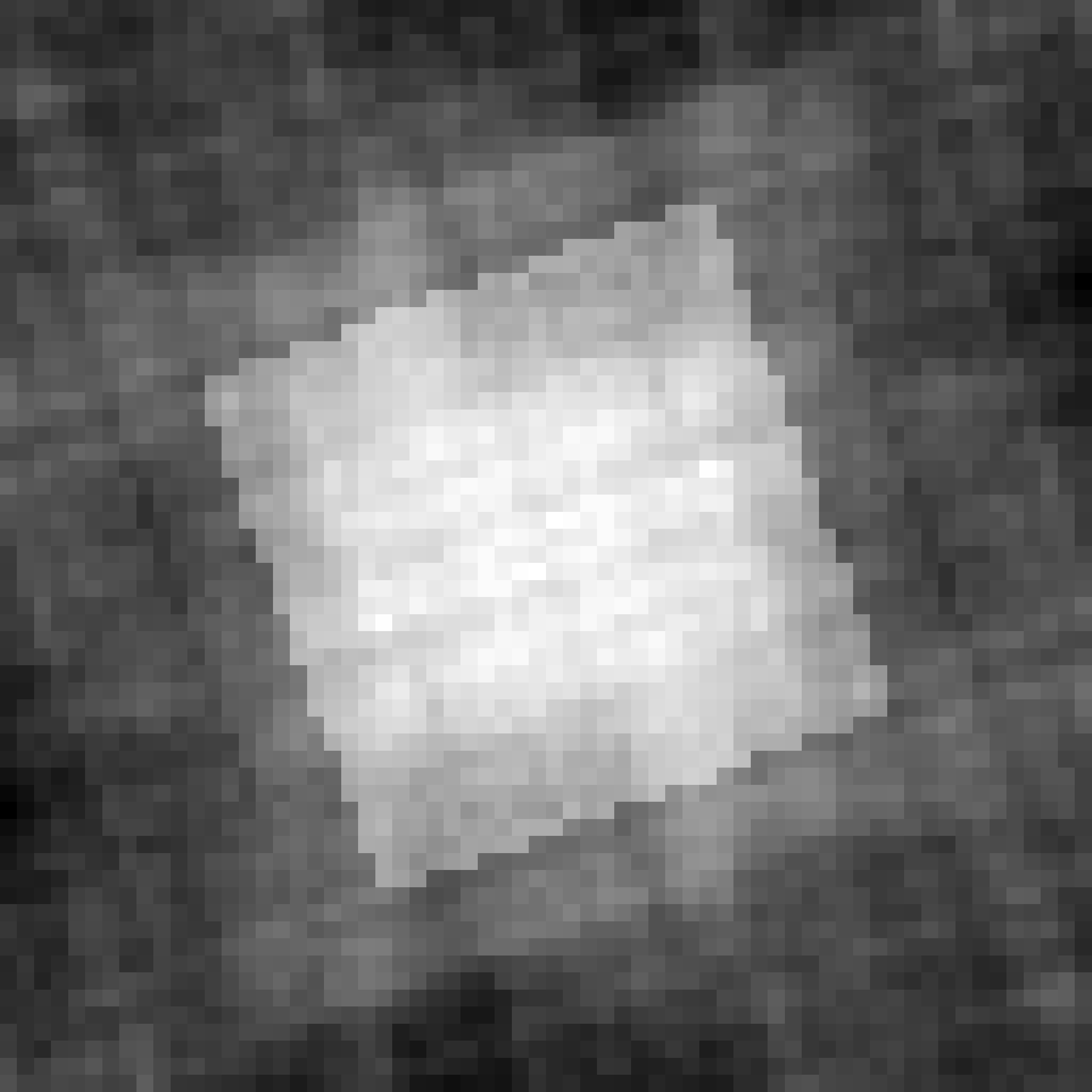}
        \caption{PSNR$=43.8$dB}
        \label{fig:square_DB_tikhonov}
    \end{subfigure}%
    \begin{subfigure}[t]{0.125\textwidth}
        \centering
        \includegraphics[width=\textwidth]{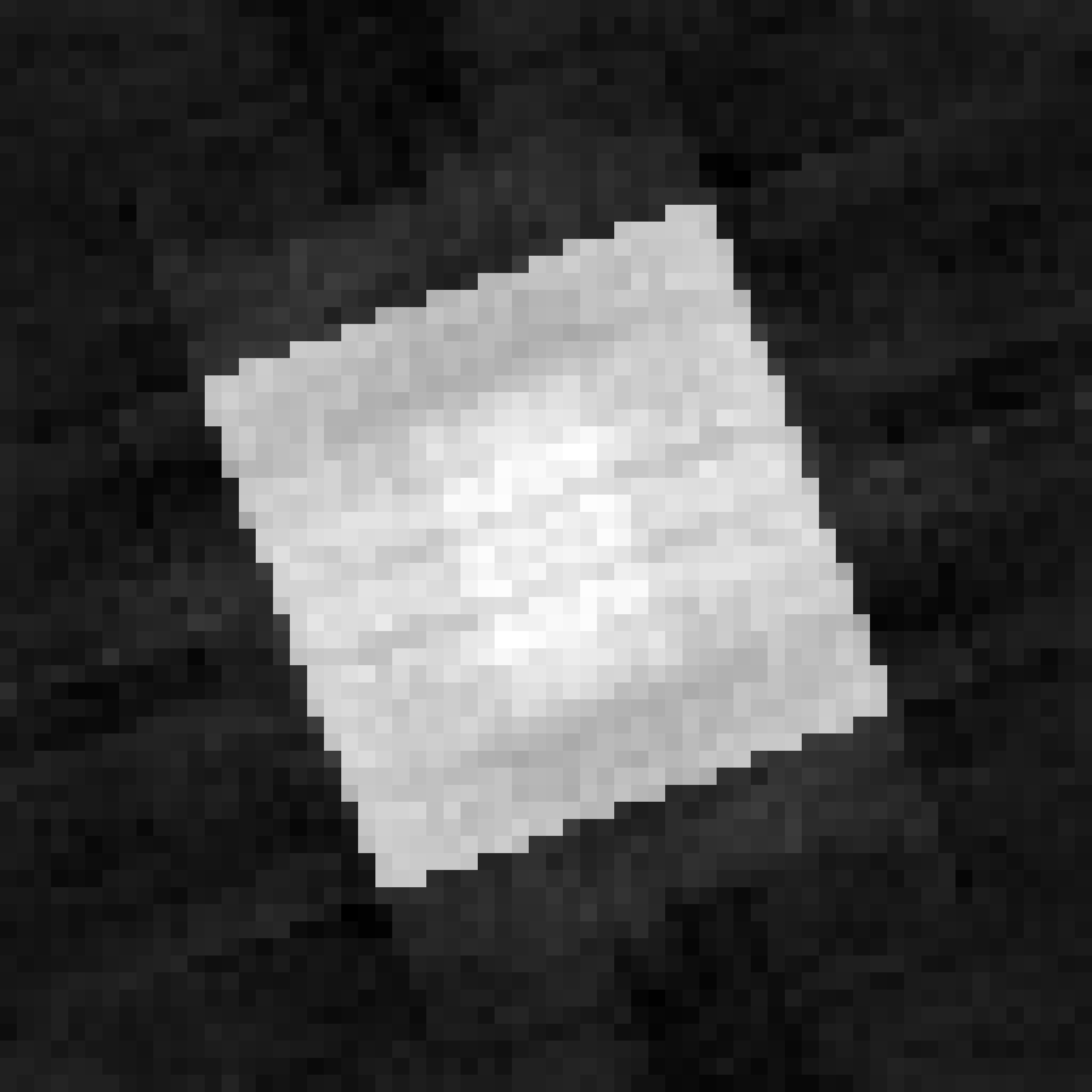}
        \caption{PSNR$=53.0$dB}
        \label{fig:square_optim_tikhonov}
    \end{subfigure}%
    \begin{subfigure}[t]{0.125\textwidth}
        \centering
        \includegraphics[width=\textwidth]{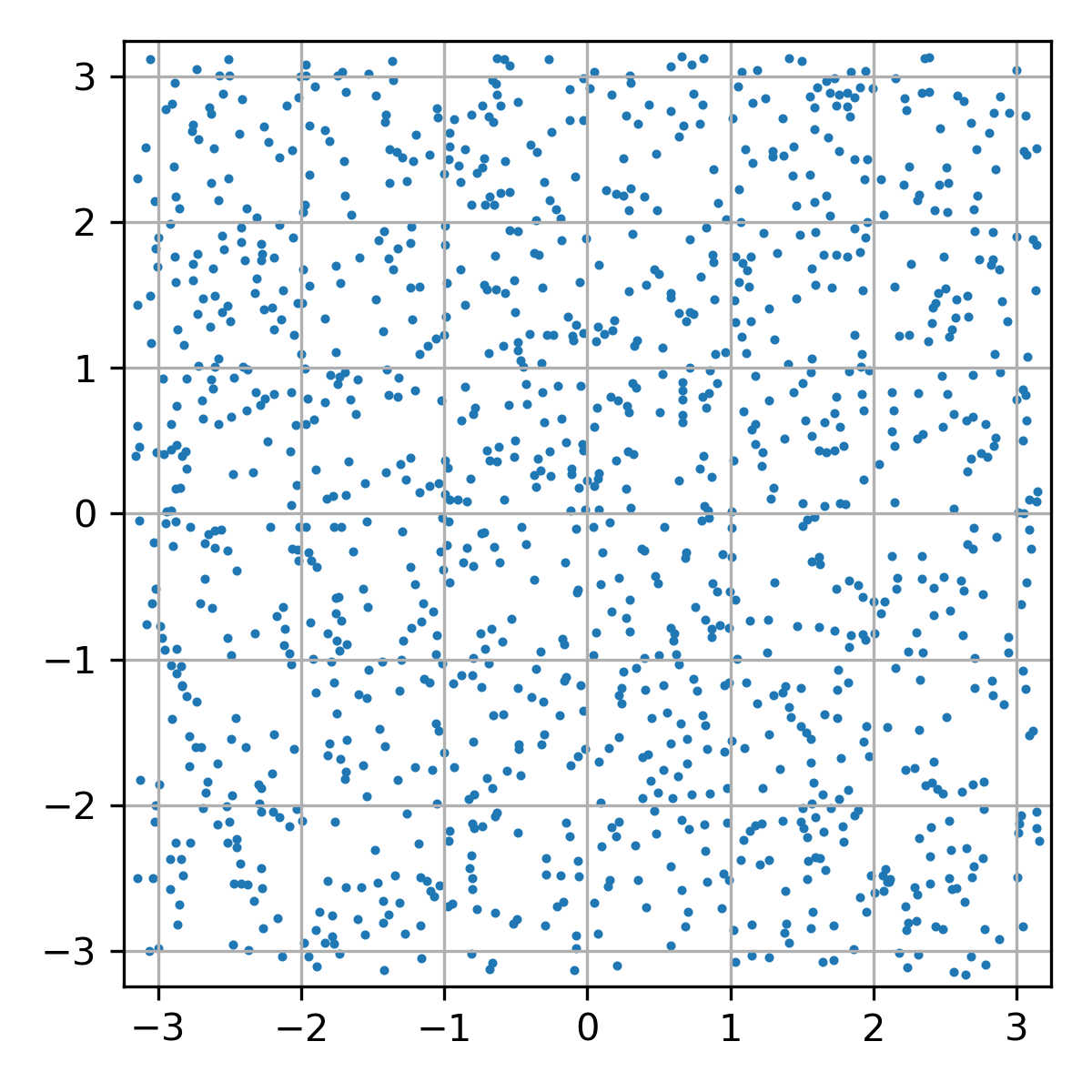}
        \caption{ }
        \label{fig:square_xi_optim_tikhonov}
    \end{subfigure}
    \caption{Reconstructed images with Tikhonov reconstructor \eqref{eq:reconstructor_tikhonov} for LF (\ref{fig:square_BF_tikhonov}), VDS (\ref{fig:square_DB_tikhonov}), OSP (\ref{fig:square_optim_tikhonov}) given in (\ref{fig:square_xi_optim_tikhonov}).}
\end{figure}

\begin{figure}[t!]
    \centering
    \begin{subfigure}[t]{0.125\textwidth}
        \centering
        \includegraphics[width=\textwidth]{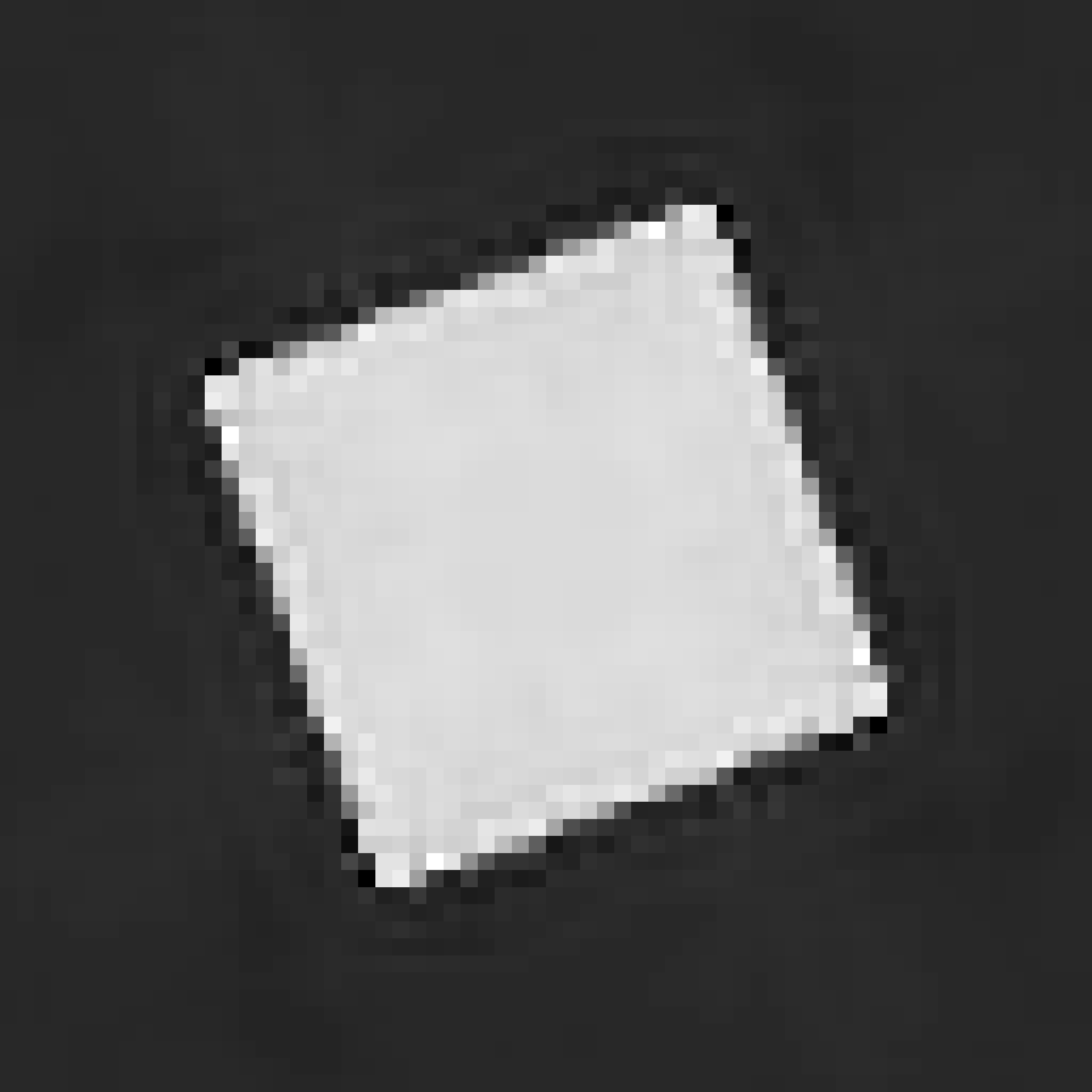}
        \caption{PSNR$=56.5$dB}
        \label{fig:square_BF_nonLin}
    \end{subfigure}%
    \begin{subfigure}[t]{0.125\textwidth}
        \centering
        \includegraphics[width=\textwidth]{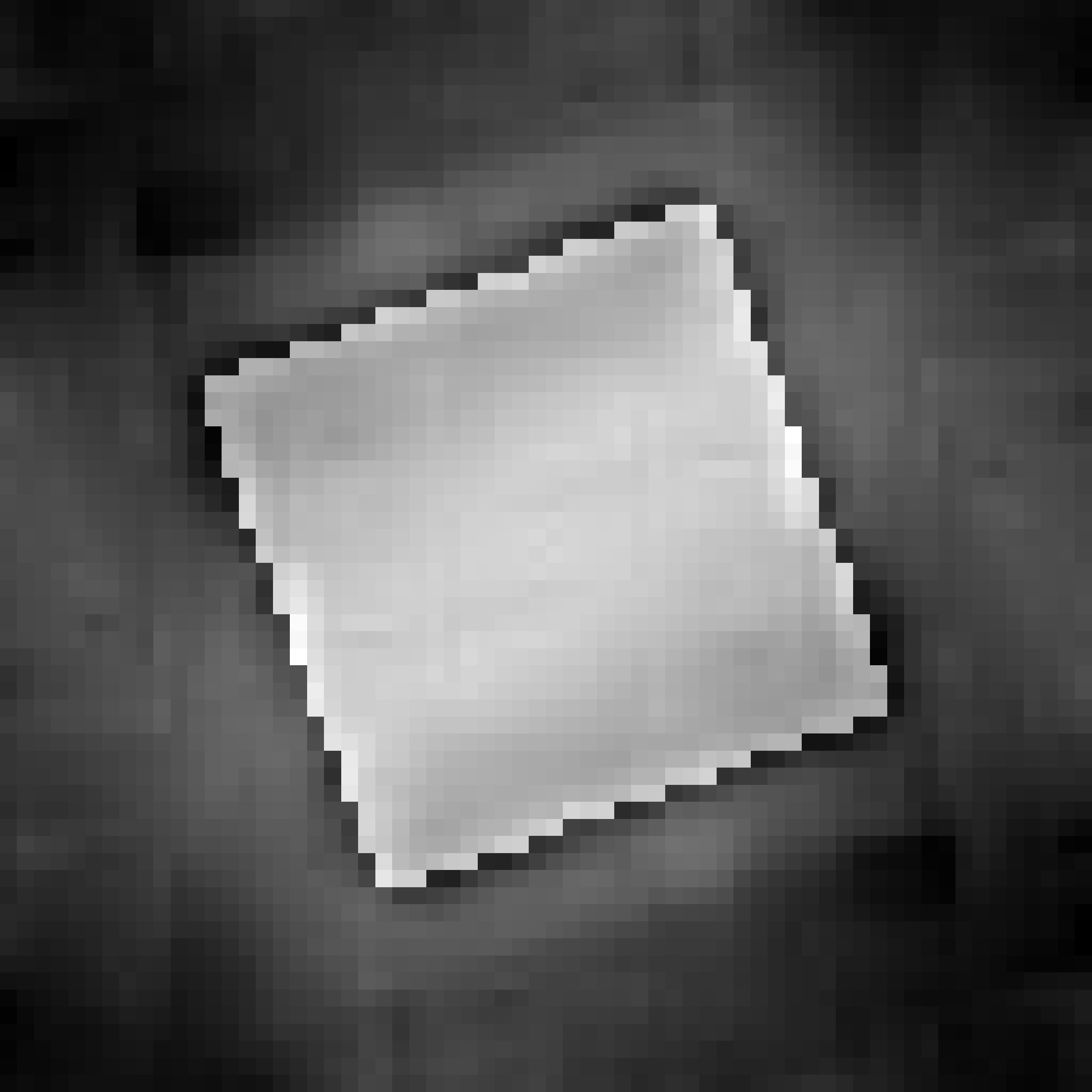}
        \caption{PSNR$=45.0$dB}
        \label{fig:square_DB_nonLin}
    \end{subfigure}%
    \begin{subfigure}[t]{0.125\textwidth}
        \centering
        \includegraphics[width=\textwidth]{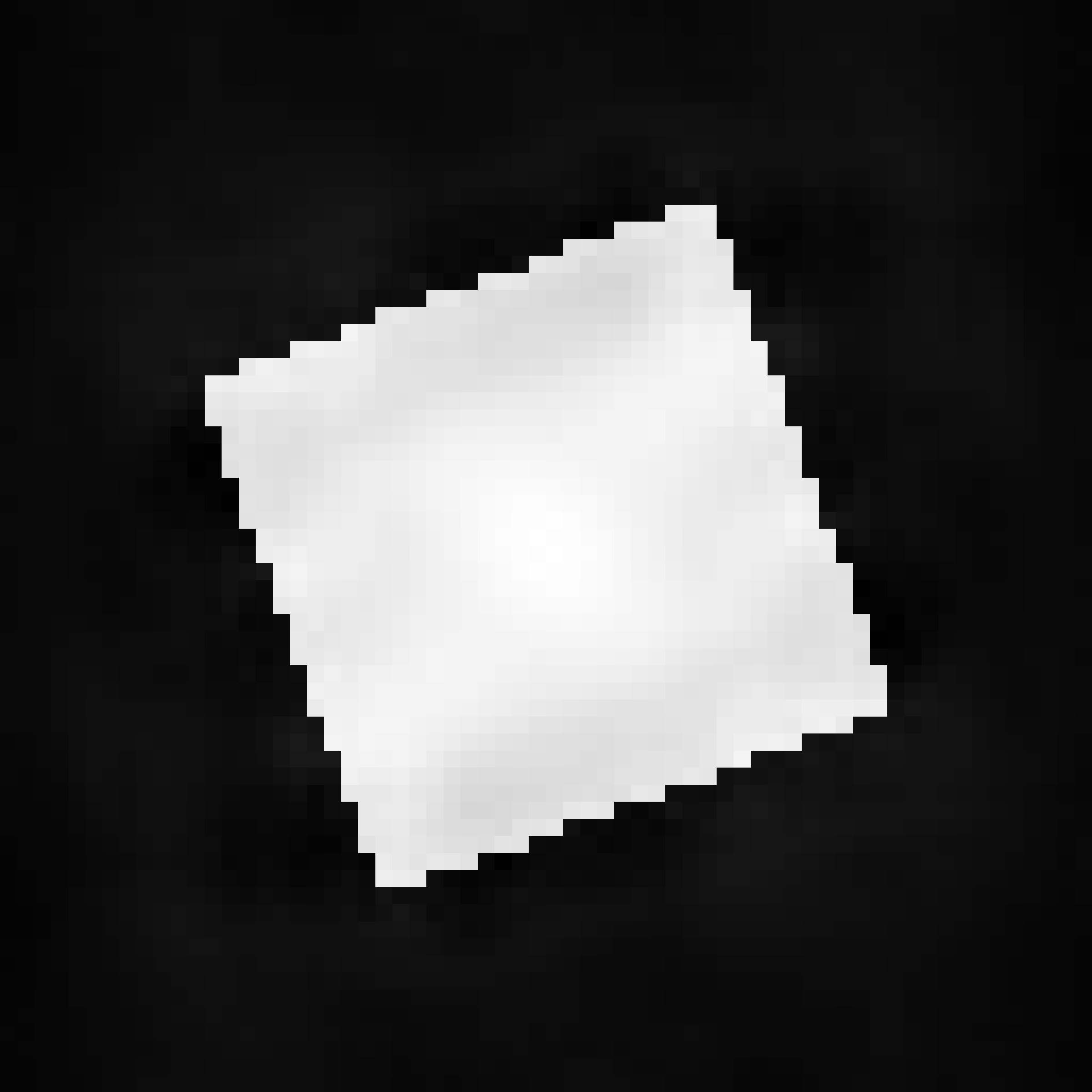}
        \caption{PSNR$=62.2$dB}
        \label{fig:square_optim_nonLin}
    \end{subfigure}%
    \begin{subfigure}[t]{0.125\textwidth}
        \centering
        \includegraphics[width=\textwidth]{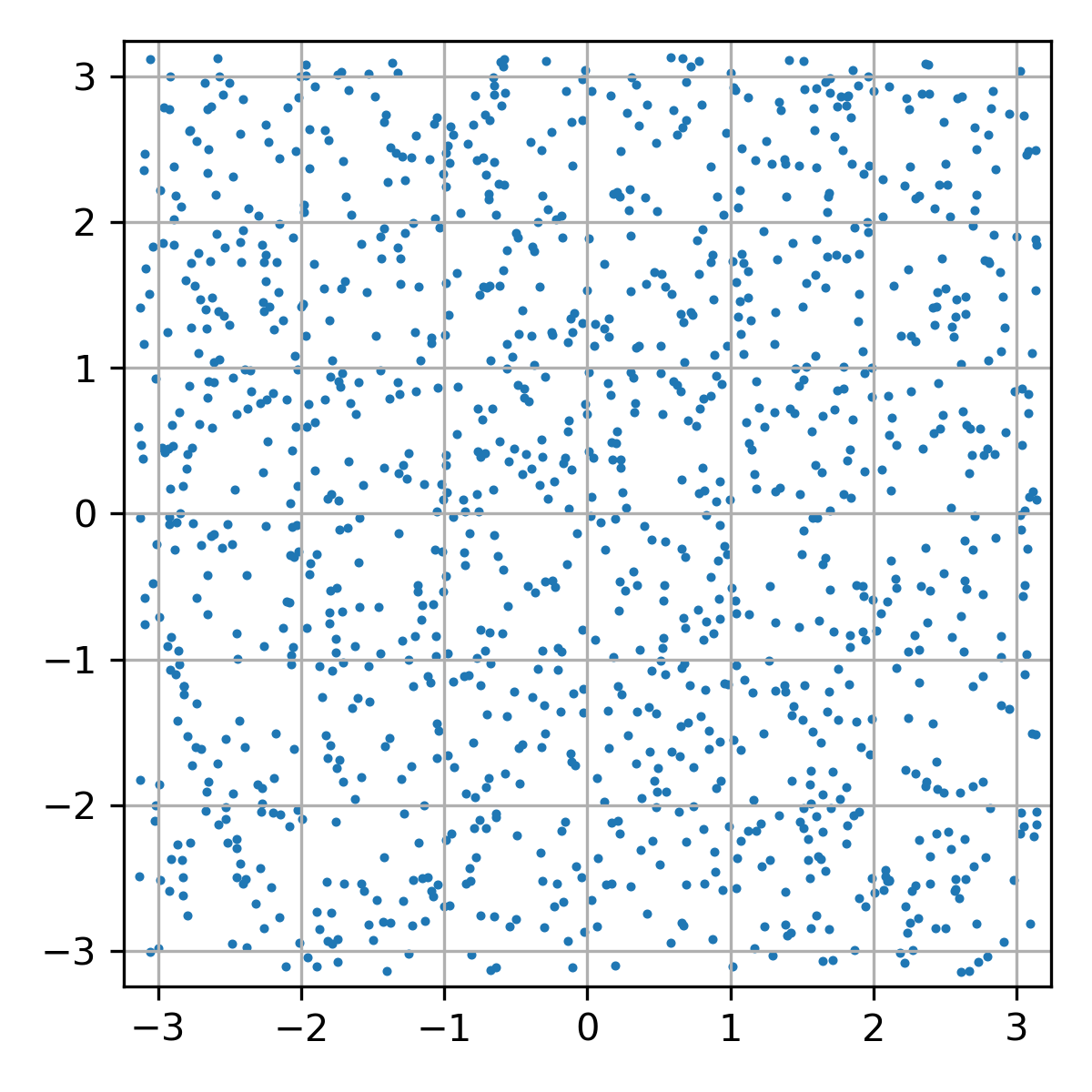}
        \caption{ }
        \label{fig:square_xi_optim_nonLin}
    \end{subfigure}
    \caption{Reconstructed images with nonlinear reconstructor \eqref{eq:reconstructor_nonlinear} for LF (\ref{fig:square_BF_nonLin}), VDS (\ref{fig:square_DB_nonLin}), OSP (\ref{fig:square_optim_nonLin}) given in (\ref{fig:square_xi_optim_nonLin}).}
\end{figure}

\begin{figure}[t!]
    \centering
    \begin{subfigure}[t]{0.125\textwidth}
        \centering
        \includegraphics[width=\textwidth]{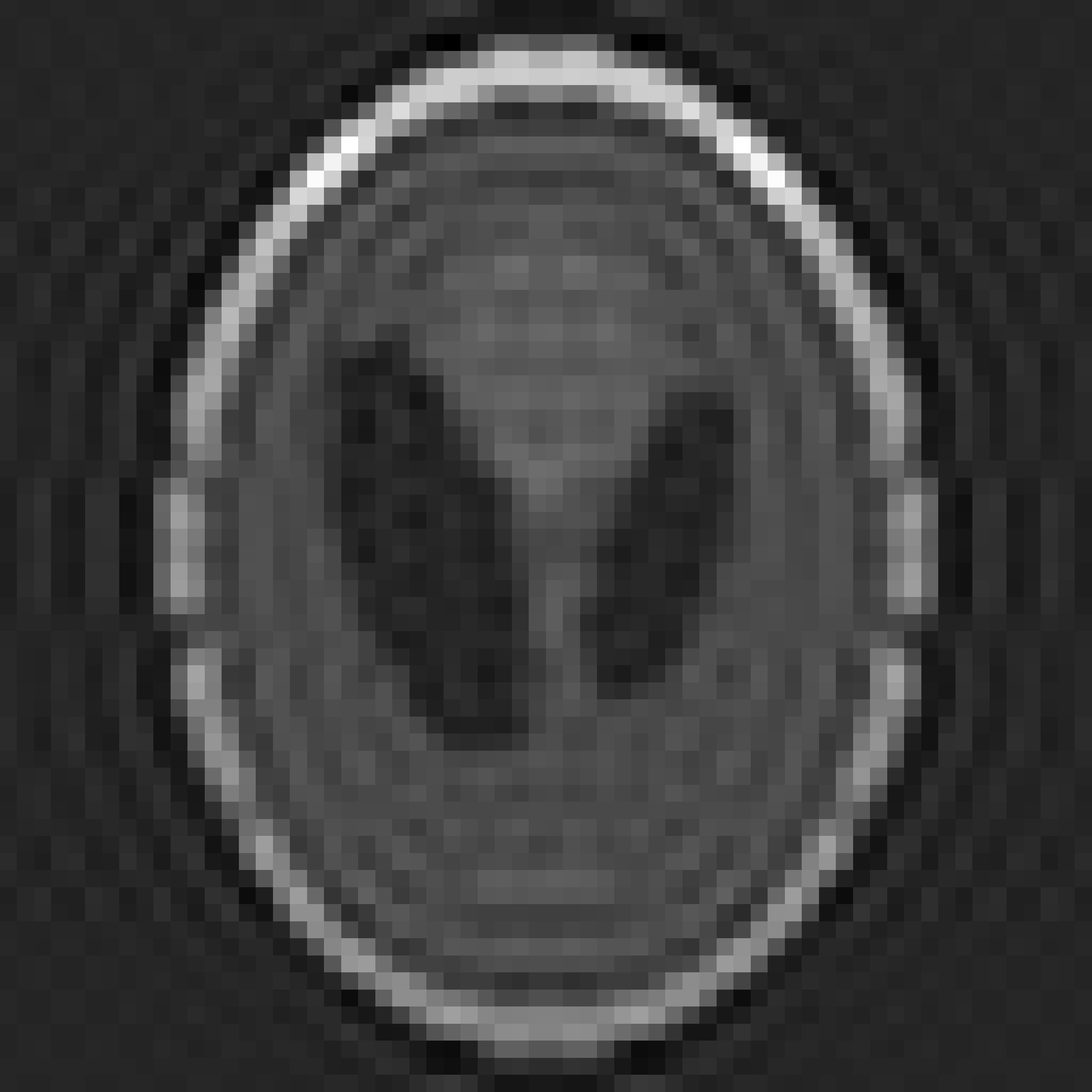}
        \caption{PSNR$=43.3$dB}
        \label{fig:phantom_BF_tikhonov}
    \end{subfigure}%
    \begin{subfigure}[t]{0.125\textwidth}
        \centering
        \includegraphics[width=\textwidth]{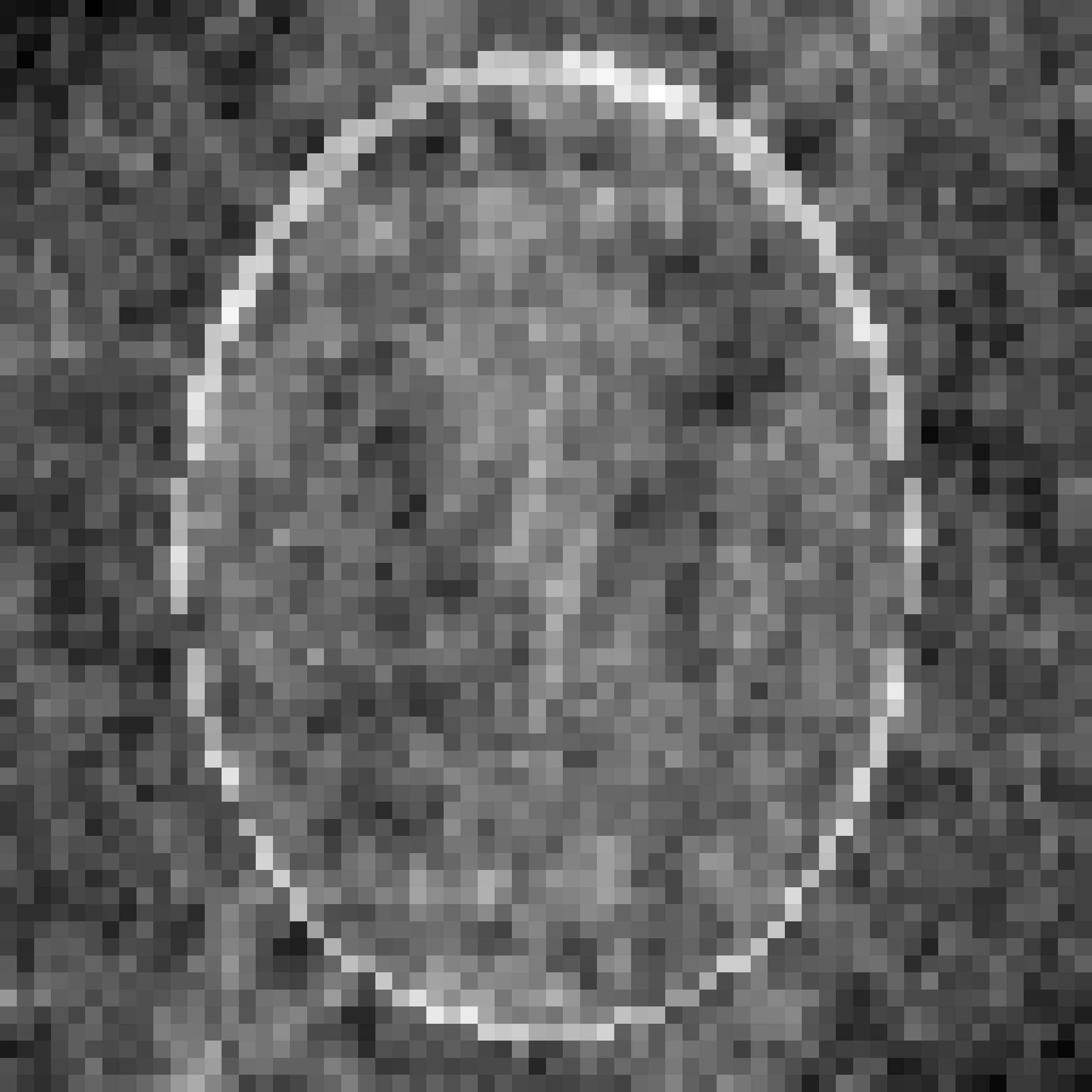}
        \caption{PSNR$=40.0$dB}
        \label{fig:phantom_DB_tikhonov}
    \end{subfigure}%
    \begin{subfigure}[t]{0.125\textwidth}
        \centering
        \includegraphics[width=\textwidth]{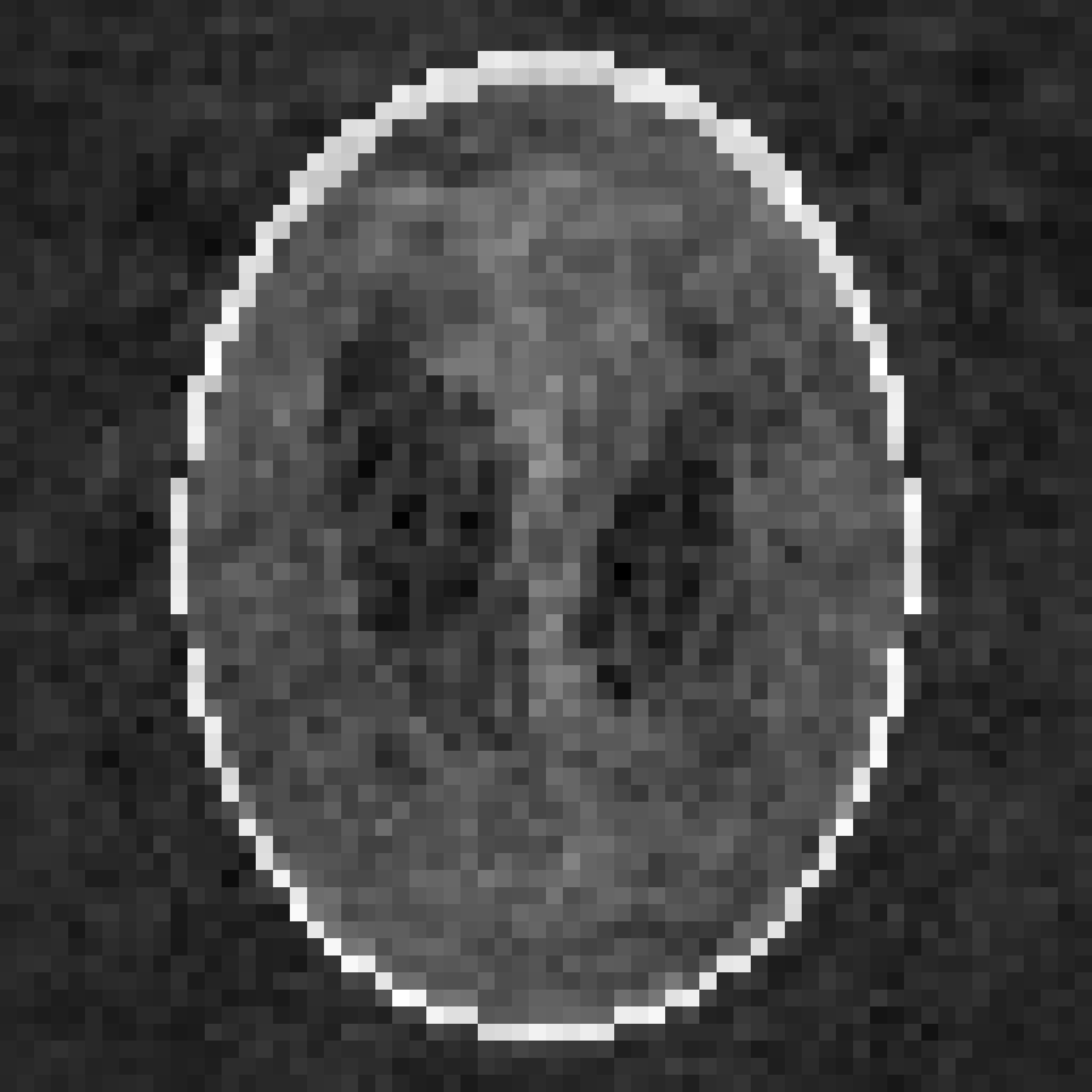}
        \caption{PSNR$=47.3$dB}
        \label{fig:phantom_optim_tikhonov}
    \end{subfigure}%
    \begin{subfigure}[t]{0.125\textwidth}
        \centering
        \includegraphics[width=\textwidth]{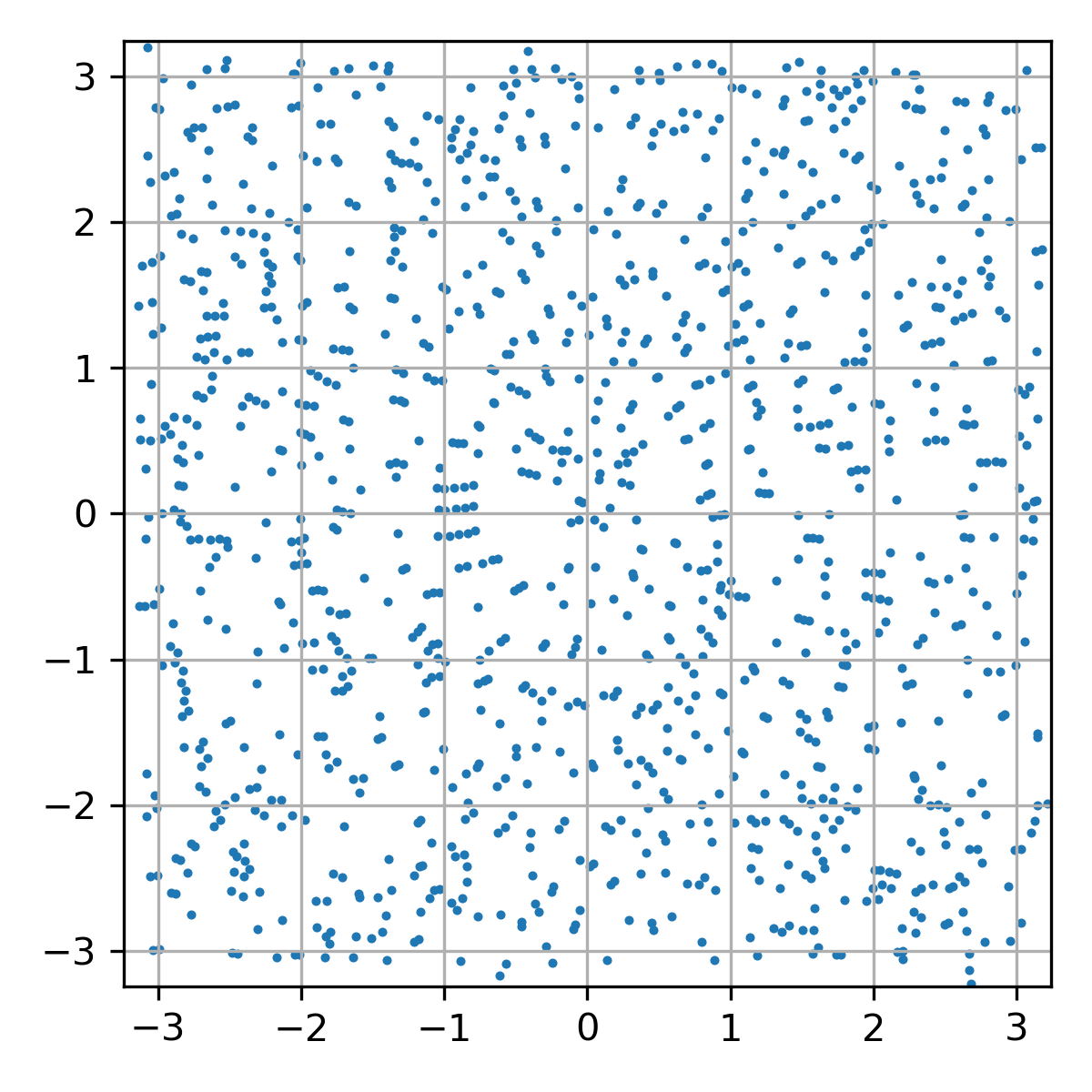}
        \caption{ }
        \label{fig:phantom_xi_optim_tikhonov}
    \end{subfigure}
    \caption{Reconstructed images with Tikhonov reconstructor \eqref{eq:reconstructor_tikhonov} for LF (\ref{fig:phantom_BF_tikhonov}), VDS (\ref{fig:phantom_DB_tikhonov}), OSP (\ref{fig:phantom_optim_tikhonov}) given in (\ref{fig:phantom_xi_optim_tikhonov}).}
\end{figure}

\begin{figure}[t!]
    \centering
    \begin{subfigure}[t]{0.125\textwidth}
        \centering
        \includegraphics[width=\textwidth]{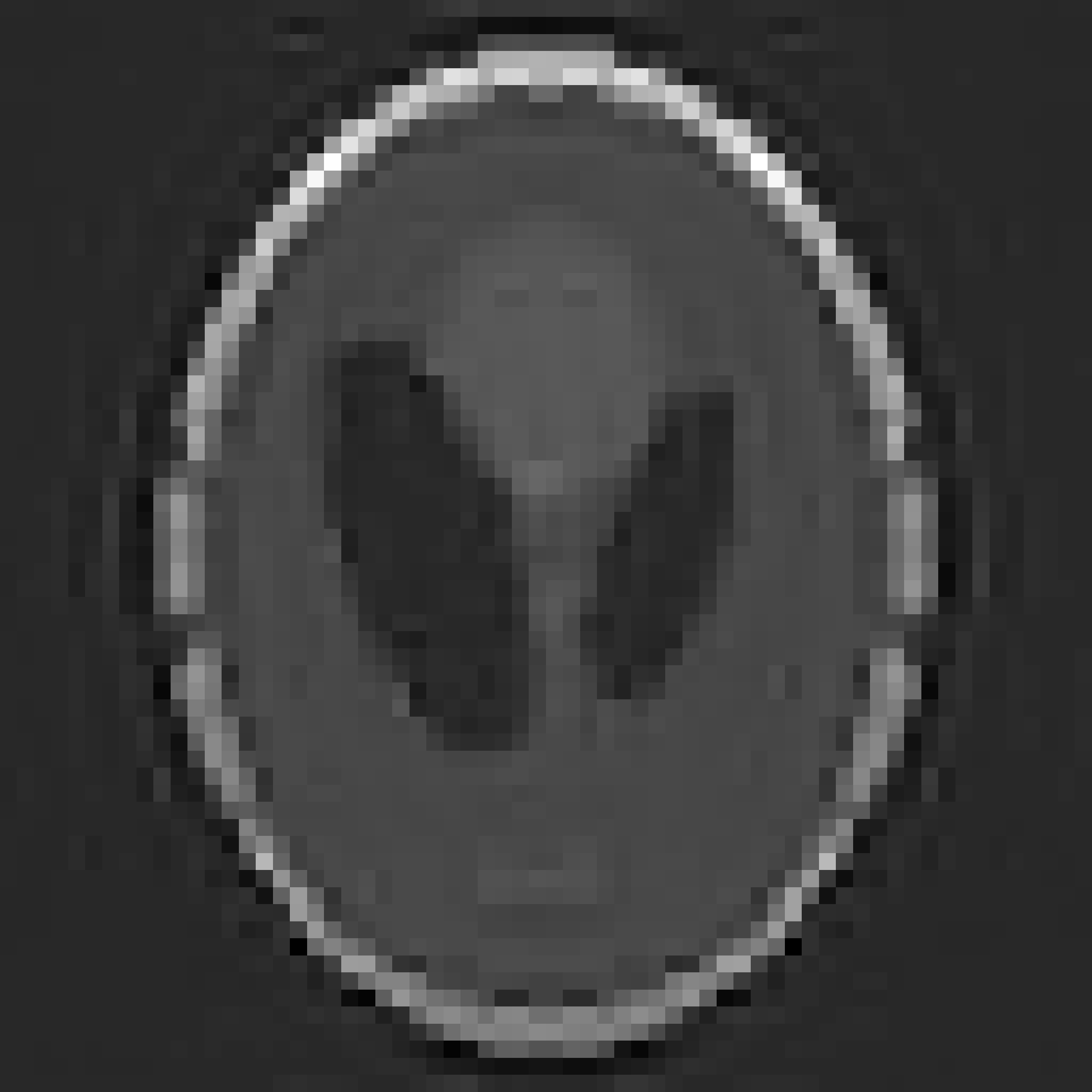}
        \caption{PSNR$=44.3$dB}
        \label{fig:phantom_BF_nonLin}
    \end{subfigure}%
    \begin{subfigure}[t]{0.125\textwidth}
        \centering
        \includegraphics[width=\textwidth]{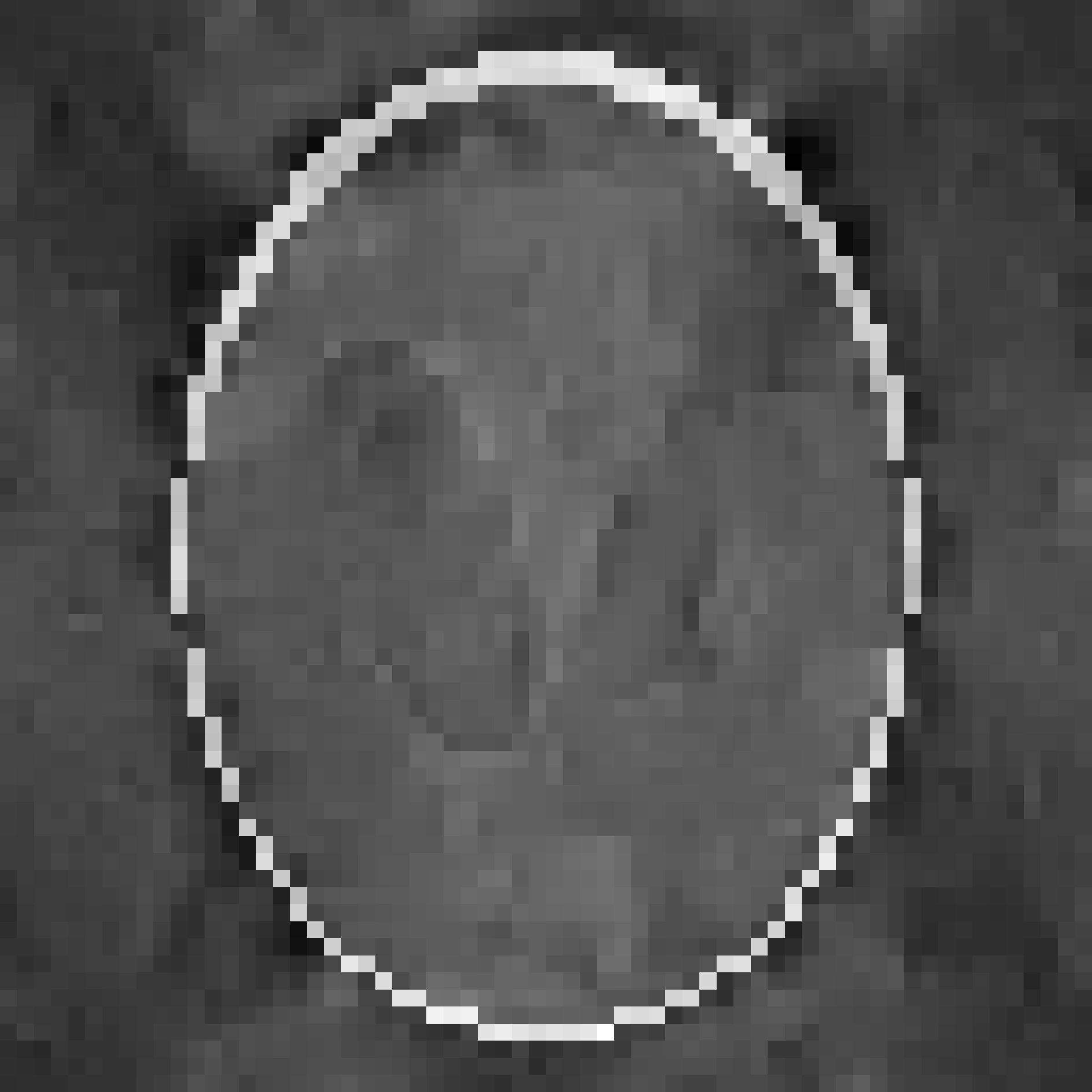}
        \caption{PSNR$=44.5$dB}
        \label{fig:phantom_DB_nonLin}
    \end{subfigure}%
    \begin{subfigure}[t]{0.125\textwidth}
        \centering
        \includegraphics[width=\textwidth]{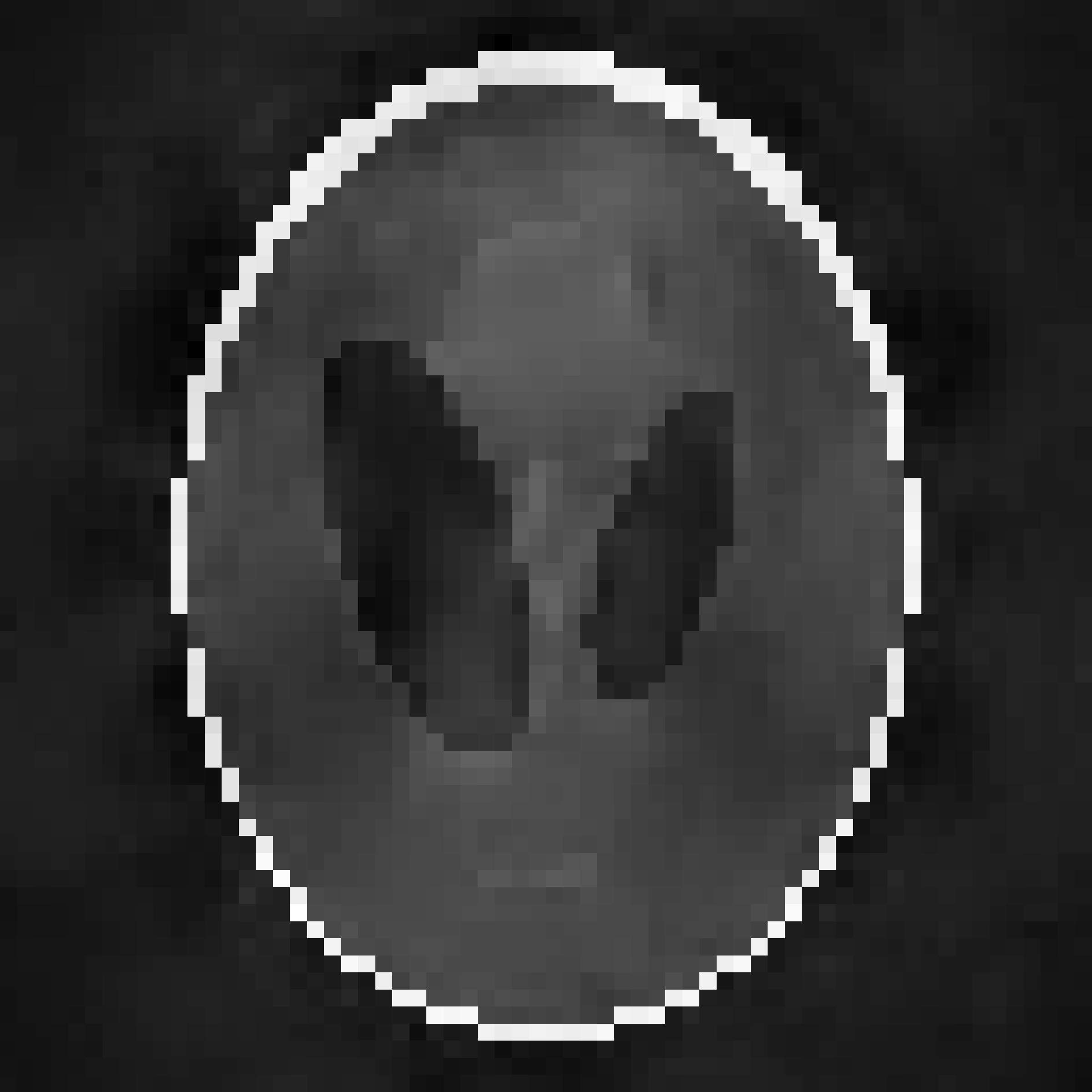}
        \caption{PSNR$=51.8$dB}
        \label{fig:phantom_optim_nonLin}
    \end{subfigure}%
    \begin{subfigure}[t]{0.125\textwidth}
        \centering
        \includegraphics[width=\textwidth]{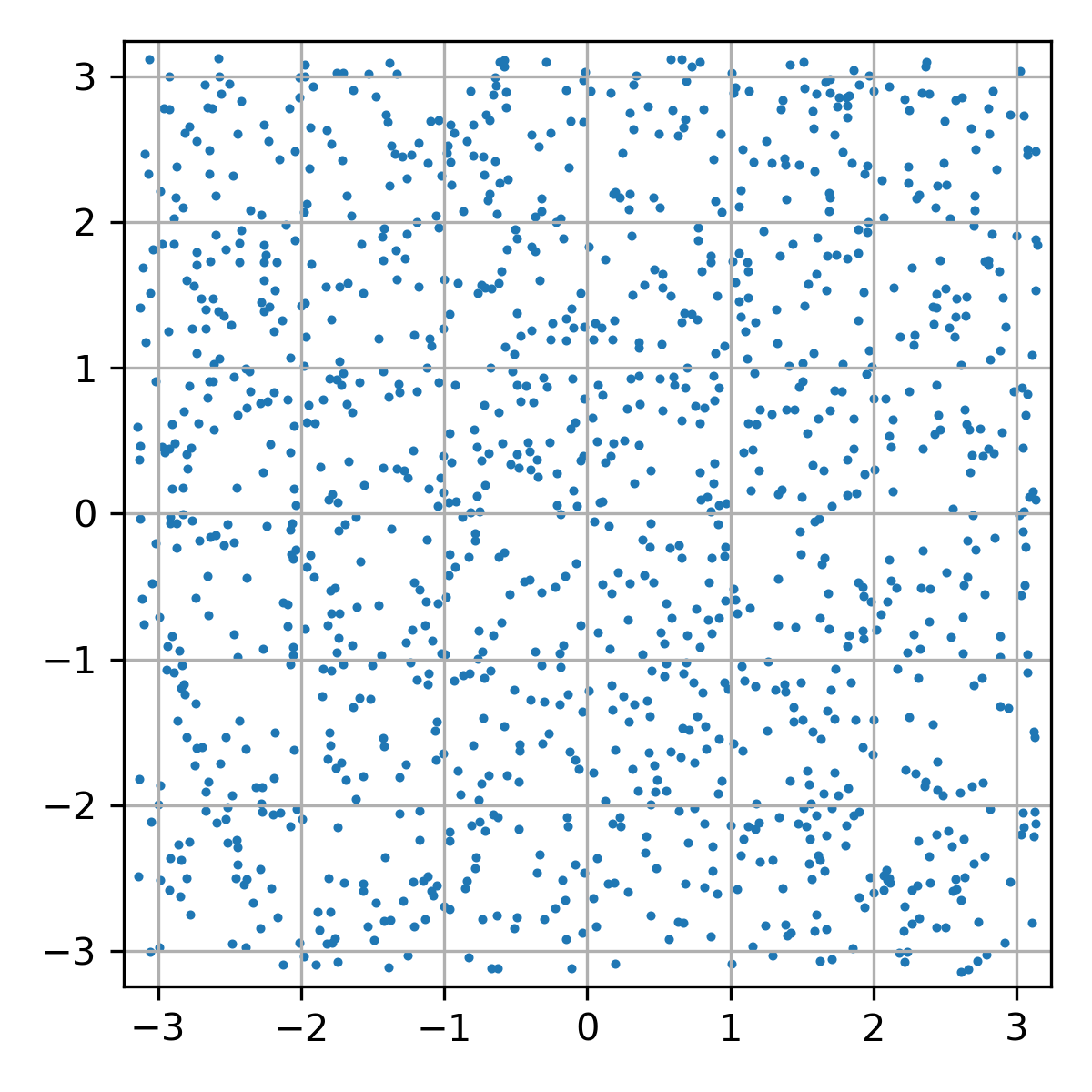}
        \caption{ }
        \label{fig:phantom_xi_optim_nonLin}
    \end{subfigure}
    \caption{Reconstructed images with nonlinear reconstructor \eqref{eq:reconstructor_nonlinear} for LF (\ref{fig:phantom_BF_nonLin}), VDS (\ref{fig:phantom_DB_nonLin}), OSP (\ref{fig:phantom_optim_nonLin}) given in (\ref{fig:phantom_xi_optim_nonLin}).}
\end{figure}

\vspace{-0.1cm}
\section*{Acknowledgments}
This work was supported by the ANR Optimization on Measures Spaces and of ANR-3IA Artificial and Natural Intelligence Toulouse Institute.

\clearpage
\bibliographystyle{plain}
\bibliography{biblio}

\end{document}